\documentclass{article}
\usepackage{iclr2027_conference,times}
\usepackage[T1]{fontenc}
\usepackage{amsmath,amssymb,booktabs,graphicx,tabularx,array,multirow}
\usepackage{microtype,placeins,capt-of}
\usepackage{hyperref,url}
\usepackage{xcolor}
\hypersetup{hidelinks,
  pdftitle={PREFER: From Player Experience to Personalized Game Refinement},
  pdfauthor={Qi Chai, Zhang Zheng, Lu Yang, Zixun Sun, Hao Wang}}

\newcommand{\method}{\textsc{Prefer}}
\graphicspath{{figures/}}
\title{PREFER: From Player Experience\\to Personalized Game Refinement}
\author{Qi Chai\thanks{Equal contribution.}\quad
Zhang Zheng\footnotemark[1]\quad
Lu Yang\quad Zixun Sun\quad
Hao Wang\thanks{Corresponding author.}}
\iclrfinalcopy
\begin{document}
\maketitle
\fancyhead{}
\renewcommand{\headrulewidth}{0pt}
\begin{abstract}
Understanding how large language models (LLMs) evaluate gameplay experience is important for using their feedback to guide game design. However, overall ratings alone provide limited insight into which aspects of experience are associated with these judgments and how they can inform concrete game changes. We introduce \method{}, a framework that uses a shared set of experience dimensions adapted from established psychometric questionnaires to benchmark how models evaluate gameplay and to guide persona-specific game refinement. Through these dimensions, we characterize the perspectives models offer as virtual playtesters by examining how their overall assessments relate to specific aspects of experience and to the gameplay they produce. To examine how the same dimensions can guide refinement for players with different preferences, we use personas to simulate distinct player profiles. For each persona, \method{} uses experience-preference relationships to select refinement objectives, revises game rules and their implementations, and playtests candidate games to assess experience changes and predicted preference gains. Experiments with five personas show that \method{} achieves {9.2\%} higher mean post-play preference ratings than {the strongest evaluated baseline}.
\end{abstract}

\section{Introduction}
\vspace{-5pt}
Large language models (LLMs) are increasingly used in game development, supporting the generation of playable prototypes and the evaluation of player experience \citep{sun2024factorsim,lankes2023game}. Gameplay experience involves multiple aspects, such as challenge, mastery, and immersion \citep{sweetser2005gameflow,vanden2020development}. Understanding how these aspects relate to models' overall judgments is therefore important for characterizing the perspectives they offer as virtual playtesters and for using their feedback to guide game refinement.

Established player-experience instruments, such as PXI and CORGIS, provide structured dimensions for describing functional, psychosocial, and challenge-related aspects of gameplay \citep{vanden2020development,denisova2020measuring}. Building on such instruments, \citet{dutta2026mining} map player reviews to experience constructs and examine their associations with review scores. In parallel, MeepleLM generates ratings and critiques from game rulebooks under different player personas, illustrating how LLMs can provide feedback from multiple player perspectives \citep{li2026meeplelm}.

Building on these advances, we ask how different LLMs evaluate the experience reflected in actual gameplay. An overall rating alone does not reveal whether models agree on specific aspects of an experience, or how those assessments relate to their overall judgments. Characterizing their perspectives therefore calls for comparing both levels of evaluation and examining how these judgments relate to the gameplay the models produce. Beyond this analysis, a further question is whether the same experience dimensions can guide game refinement for players with different preferences, by specifying which aspects to change and in which direction.

To address these questions, we introduce \method{}, a framework that uses shared experience dimensions to benchmark how models evaluate gameplay and to guide persona-specific game refinement. We adapt experience dimensions from established psychometric questionnaires to assess gameplay trajectories, collecting overall ratings separately from dimension-level assessments. Different models evaluate the same trajectories, allowing us to examine how their overall judgments relate to specific aspects of experience and how their assessments agree or differ. We further compare their evaluations of trajectories produced through their own play and that of other models, and examine agreement in game-level rankings.

However, players differ in what they value in a game \citep{tondello2019don}. To better simulate diverse real-world players, we follow MeepleLM \citep{li2026meeplelm} in using personas to represent distinct player profiles. For each persona, \method{} relates experience dimensions to overall preference to select which aspects to change and in which direction. These objectives guide rule revisions, which are translated into function-level contracts that specify required behavior and guide implementation checks and repair. Candidate games are then playtested under the target persona, and acceptance requires both the intended experience changes and a positive predicted preference gain. Feedback from unsuccessful candidates guides further rule revisions.

We evaluate \method{} on {familiar} card games and their rule variants. The model benchmark identifies shared patterns in game rankings alongside differences in how experience dimensions relate to overall ratings. Across five personas, \method{} achieves {9.2\%} higher mean post-play preference ratings than the strongest baseline evaluated. A separate implementation study also shows a higher success rate in producing runnable games and higher rule-consistency scores than baseline.

Our contributions are threefold:
\vspace{-5pt}

\begin{itemize}
\item We introduce \method{}, a framework that adapts experience dimensions from psychometric questionnaires to gameplay trajectories, providing a shared basis for model evaluation and persona-specific game refinement.
\item We establish a model benchmark that characterizes the perspectives LLMs offer as virtual playtesters by comparing their experience judgments, overall ratings, and game rankings, and relating these evaluations to the gameplay they produce.
\item We develop a persona-specific game refinement method that uses experience-preference relationships to select modification targets, guides rule revisions with playtesting feedback, and supports implementation through function-level contracts and execution checks.
\end{itemize}

\begin{figure}[t]
\centering\includegraphics[width=0.9\linewidth]{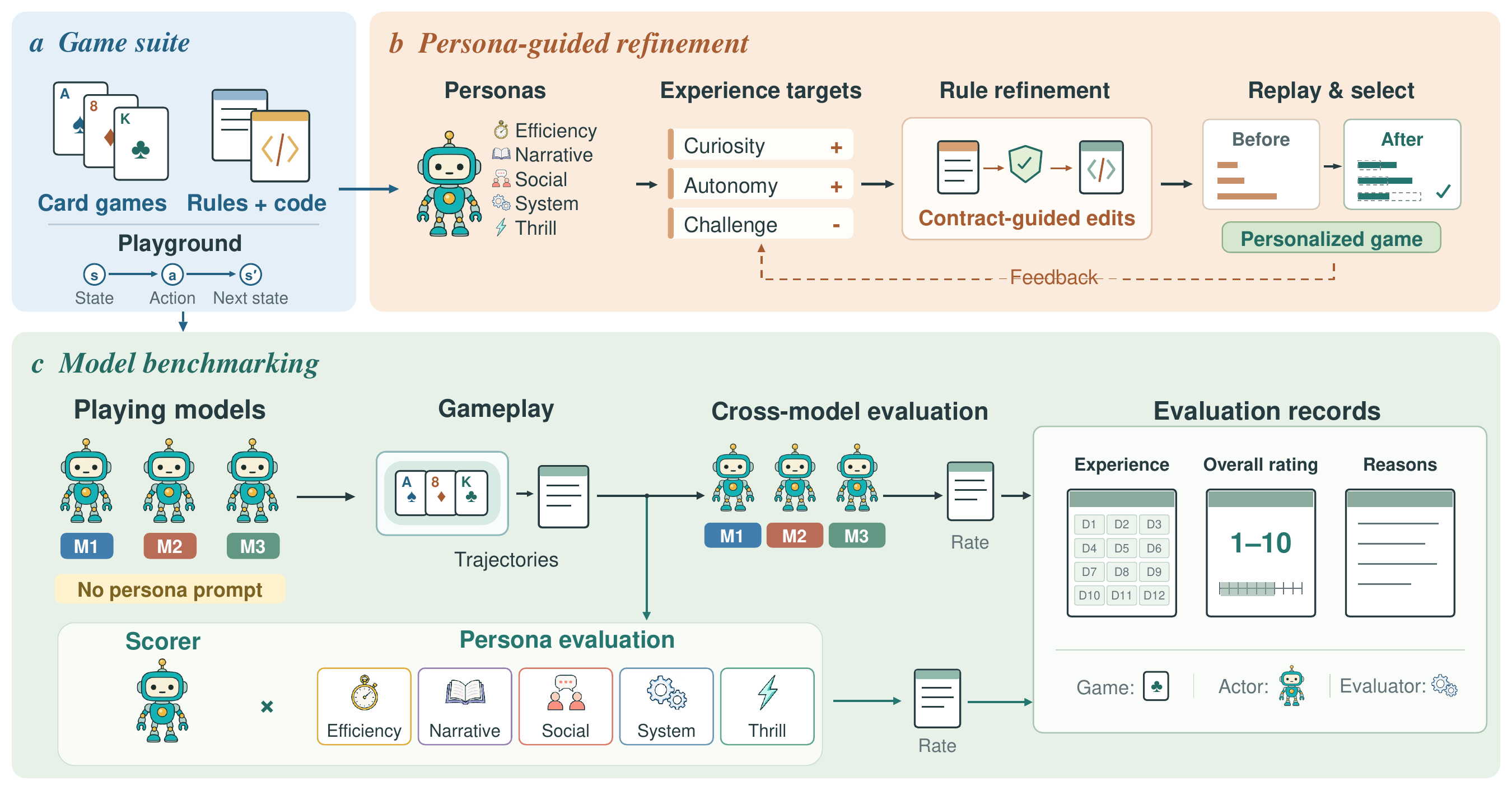}

\caption{\textbf{Overview of \method{}.} (a) The game suite provides rules, code, and a playground. (b) Persona preferences guide rule refinement, implementation, and playtesting. (c) Models and persona-based evaluators assess the same gameplay trajectories using shared experience dimensions and overall ratings.}
\label{fig:overview}
\vspace{-15pt}
\end{figure}
\vspace{-10pt}
\section{Related Work}

\paragraph{Game-Based Agent Evaluation.}
Games let researchers observe how agents follow rules, choose actions, and respond to other players over several turns. clembench tests instruction following and task success through dialogue games \citep{kranti2023clembench}. GameBench studies strategic reasoning across nine games \citep{costarelli2024gamebench}. TextArena offers competitive text games that involve skills such as negotiation and deception \citep{guertler2025textarena}. Our benchmark uses shared experience measures to compare models without persona prompts against reference players given fixed persona prompts. We compare their evaluation profiles on shared gameplay evidence and relate these judgments to the play the models produce.
\vspace{-8pt}
\paragraph{Personas and Player Experience.}
Persona prompts ask an agent to follow a given role or set of preferences. TRAIT measures model personality through multiple-choice questions about everyday situations \citep{lee2025llms}. PERIL studies how persona prompts affect game strategies. It maps a model's answers to personality questions into strategy parameters \citep{licato2025persona}. MeepleLM uses player personas to rate games and explain their judgments from rule descriptions \citep{li2026meeplelm}. PXI and CORGIS offer scales for player experience and perceived challenge \citep{vanden2020development,denisova2020measuring}. We use these scales to describe gameplay. Personas guide our game revisions and provide reference players for model comparison.
\vspace{-8pt}
\paragraph{Game Refinement and Verification.}
Game refinement requires both useful feedback and code that follows the revised rules. Self-Refine uses model feedback to improve generated outputs \citep{madaan2023self} and systems~\citep{yuksekgonul2025optimizing,202609.0591}. Cardiverse creates card game variants and playable implementations \citep{li2025cardiverse}. Fly, Fail, Fix uses play records from a reinforcement learning agent to guide changes to game settings \citep{zook2025fly}. For code repair, LDB checks execution step by step to locate errors \citep{zhong2024debug}. PolyVer uses function contracts to verify programs written in different languages \citep{chen2025polyver}. In our method, a target player's experience guides rule changes. Fixed function contracts guide code repair, and playtesting checks the resulting experience.
\vspace{-8pt}
\section{Method}
\vspace{-5pt}
\label{sec:method}

\method{} uses shared experience dimensions to study how LLMs evaluate gameplay and to guide game refinement for a target persona. We first describe the experience assessment. We then explain how persona preferences guide rule changes, how function contracts guide their implementation, and how playtesting provides feedback for further revision.
\vspace{-5pt}
\subsection{Experience Assessment}
\label{sec:experience-assessment}
\vspace{-5pt}
We describe gameplay through selected dimensions from PXI \citep{vanden2020development} and CORGIS \citep{denisova2020measuring}. These dimensions cover how a game supports and engages the player, as well as the demands of play. For example, challenge fit measures whether difficulty suits the player. Cognitive challenge measures the reasoning and planning required during play. We keep the dimension definitions and scoring scales fixed across models and personas. Appendix~\ref{app:representation} gives the full definitions and scoring details.

Let $R$ denote the rulebook and $\tau$ a completed gameplay trajectory. The trajectory records the focal player's observations, available actions, chosen actions, and outcomes. An LLM evaluator uses this evidence to rate each experience dimension. A separate assessment gives an overall rating of the same playthrough. We write the two assessments as
\begin{equation}
x=\mathcal{E}_m(R,\tau,p),
\qquad
y=\mathcal{S}_m(R,\tau,p),
\label{eq:assessment}
\end{equation}
where $m$ is the evaluation model, $x\in\mathbb{R}^{d}$ contains the $d$ experience scores, and $y$ is the overall rating. The optional persona $p$ specifies the player's preferences. Each experience score includes a short explanation based on the trajectory. The two assessments do not receive each other's outputs, so $y$ is not calculated from $x$.

For model comparison, we omit the persona prompt and compare judgments of the same gameplay evidence. The pairs $(x,y)$ let us examine how each model's dimension scores relate to its overall ratings. For persona-guided refinement, the player agent and both assessments use the same target persona $p$. In this setting, $y$ measures the persona's preference for the played game.
\vspace{-5pt}
\subsection{Rule Refinement}
\label{sec:target}
\vspace{-5pt}
Rule refinement starts from a game with rulebook $R$, code $C$, and a target persona $p$. We collect paired experience and preference measurements across games under this persona. A regularized linear predictor $f_p$ relates the experience vector to the preference score. We fit the predictor on other games before using it to refine the current game. Appendix~\ref{app:preference-model} gives the predictor and fitting procedure.

The predictor assigns a weight $w_{pj}$ to each experience dimension $j$. In the original scoring units, this weight estimates the preference change for a one-point increase in dimension $j$, with other dimensions fixed. We use the sign of $w_{pj}$ to choose the direction of change. A positive weight suggests increasing the dimension, and a negative weight suggests decreasing it. We denote this direction by $d_{pj}\in\{-1,+1\}$ and skip dimensions with zero weight.

Given the current experience vector $x$, we remove targets without enough room for change on the scoring scale. We rank the remaining targets by $|w_{pj}|$. This ranking compares the predicted gains from equal-sized changes in individual dimensions. It sets the search order, while playtesting checks the effects of an actual rule change.

For each selected target, an LLM proposes a revised rulebook $R'$ using $R$, $p$, and the target dimension and direction. The proposal changes a game mechanic while keeping the game objective and unaffected rules fixed. The next stage implements this rulebook and checks the resulting code.
\vspace{-5pt}
\subsection{Rule Implementation}
\label{sec:contracts}
\label{sec:repair}
\vspace{-5pt}
\noindent\textbf{Contract construction.}
We compare $R'$ with $R$ to identify changed requirements and relevant rules that must remain unchanged. Each requirement keeps a link to its source rule text. We also extract function definitions and direct calls from the code's abstract syntax tree. This information helps assign each required outcome to an owner function and identify functions that may supply its prerequisites.

Following \citet{chen2025polyver}, we express function responsibilities through assumptions and guarantees. We add provider links and timing conditions to form each contract clause
\begin{equation}
\kappa=(f,A,G,\mathcal{P},\alpha,\delta),
\label{eq:contract-clause}
\end{equation}
where $f$ is the owner function, $A$ is its required entry condition, and $G$ is its required outcome. The set $\mathcal{P}$ lists possible prerequisite providers. The trigger $\alpha$ activates the obligation, and $\delta$ sets its deadline. For arguments $u$, state $s$ before the call, state $s'$ after the call, and return value $r$, the clause requires
\begin{equation}
A(u,s)\;\Longrightarrow\;G(s,s',r).
\label{eq:contract-behavior}
\end{equation}
Both $A$ and $G$ are executable checks. We check that the plan covers the rule requirements, has valid function links, and can execute its predicates. Semantic review provides advice on whether the clauses match the rules. Once the plan is accepted, its predicates and clause identities stay fixed during code generation and repair.

\begin{figure}[t]
\centering\includegraphics[width=0.85\linewidth]{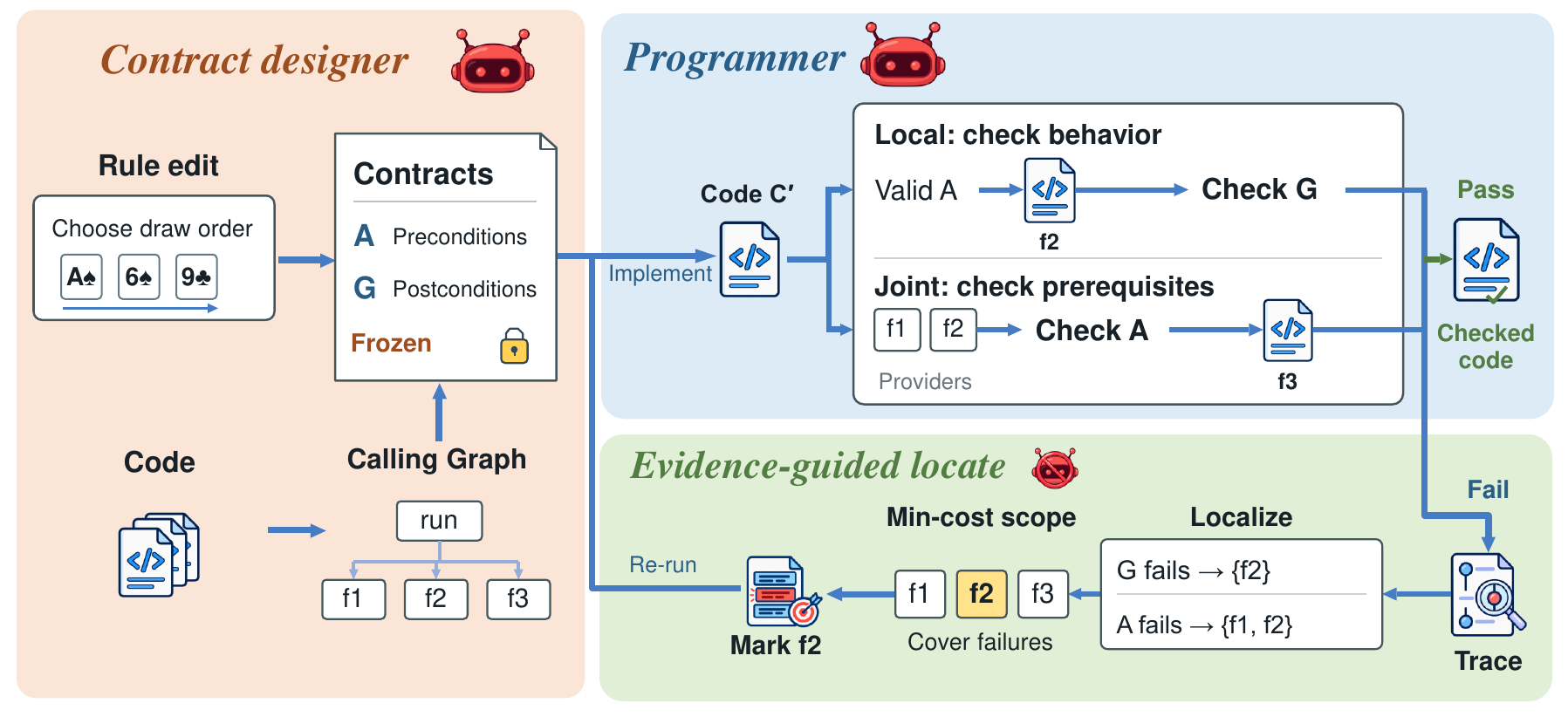}
\caption{\textbf{Contract-guided rule implementation.} Revised rules are translated into function contracts that guide code edits and execution checks. Failure traces guide the selection of functions to repair, and the updated code is rechecked under the same contracts.}
\label{fig:coordination}
\end{figure}

\noindent\textbf{Local and joint checks.}
An implementation model updates the affected function bodies using the revised rules and accepted contracts. Local checks test an owner function in isolation. Each test supplies arguments and state that satisfy $A$, independently of the candidate provider functions. It then calls the owner and tests $G$. A failed guarantee or an exception under valid entry conditions identifies an owner failure. If the test input fails $A$, the clause remains untested.

Joint checks test whether gameplay supplies the required entry conditions. A runtime monitor follows each obligation from its trigger $\alpha$. It checks $A$ when the owner is called and reports a failure if $A$ is false. A required call missing at deadline $\delta$ also causes a failure. The monitor uses provider calls observed during this period to identify possible causes. For both checks, a lack of evidence is recorded as unknown.

\noindent\textbf{Repair scope.}
We use execution feedback to select which functions to repair \citep{zhong2024debug,xia2025demystifying}. Each supported failure $i$ gives a set $E_i$ of candidate functions. A local failure gives $E_i=\{f\}$. A joint failure gives the providers supported by the execution record. We select a set that covers every failure while limiting edit cost
\begin{equation}
S^*=\arg\min_{S\subseteq\mathcal{F}}\sum_{f\in S}c_f
\quad\text{subject to}\quad
S\cap E_i\neq\varnothing\quad\text{for all }i,
\label{eq:cover}
\end{equation}
where $\mathcal{F}$ is the set of editable functions and $c_f>0$ is the number of source lines in $f$. Failures without supported candidate functions remain unresolved.

Let $\mathcal{U}$ contain the failure sets not yet covered. If $\mathcal{U}$ is nonempty, we choose a set with the fewest candidates, $E^\dagger\in\arg\min_{E\in\mathcal{U}}|E|$. The minimum remaining cost is
\begin{equation}
V(\mathcal{U})=
\begin{cases}
0, & \mathcal{U}=\varnothing,\\[2mm]
\displaystyle\min_{f\in E^\dagger}
\left[c_f+V\!\left(\{E\in\mathcal{U}\mid f\notin E\}\right)\right],
& \text{otherwise}.
\end{cases}
\label{eq:cover-recursion}
\end{equation}
Selecting $f$ covers every remaining set that contains it. We solve this recurrence by branch-and-bound search. A branch is dropped when its accumulated cost exceeds the best complete solution. A completed search returns $S^*$. Appendix~\ref{app:contracts} gives the search limits and tie-breaking rules.

The repair model receives the rules, fixed contracts, code, failure records, and selected functions. It changes only the selected function bodies and preserves their signatures. We keep a patch only if rechecking improves the results without new failures or loss of earlier evidence. Repair stops when the required checks pass, no supported repair scope remains, or the budget is exhausted. Only candidates that pass the required checks proceed to playtesting.

\begin{figure}[t]
\centering\includegraphics[width=0.85\linewidth]{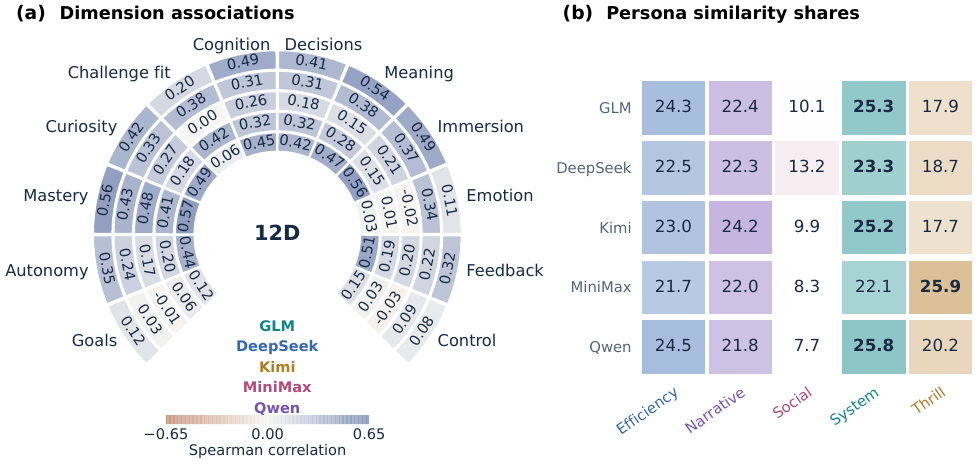}
\caption{\textbf{Experience dimensions characterize model evaluation.} (a) Associations with overall ratings reveal shared emphasis on mastery and differences across other dimensions. (b) Normalized similarity shares summarize resemblance to the five persona profiles; bold marks the largest share for each model.}
\label{fig:benchmark-experience}
\vspace{-8pt}
\end{figure}

\vspace{-5pt}
\subsection{Playtesting and Feedback}
\label{sec:selection}
\vspace{-5pt}

The player agent plays each candidate with the target persona $p$. We assess the resulting trajectory using the same dimensions and scoring procedure as for the original game. Let $x$ be the original experience vector and $x'_k$ the vector for candidate $k$. A candidate needs valid gameplay and complete experience scores before it can be accepted.

A rule change can improve the target dimension but harm other parts of the experience. We therefore check both the target change and the predicted preference gain. For candidate $k$ targeting dimension $j_k$, acceptance requires
\begin{equation}
\left\{
\begin{aligned}
d_{p j_k}\bigl(x'_{k,j_k}-x_{j_k}\bigr)&\geq\epsilon,\\
f_p(x'_k)-f_p(x)&>0,
\end{aligned}
\right.
\label{eq:admission}
\end{equation}
where $\epsilon>0$ is the minimum required change in the target dimension. The first condition checks progress in the selected direction. The second uses all experience dimensions to estimate the total preference change.

We try targets in their ranked order and generate each candidate from the same original game. A failed implementation, incomplete playtest, or rejected candidate leads to the next target. We accept the first candidate that passes all checks. If no candidate passes within the search budget, we keep the original game and enter the feedback stage.

Following \citet{madaan2023self}, we use feedback to revise a failed proposal. We select the highest-priority rejected candidate with valid gameplay and complete experience scores. The feedback includes its rules, score changes, acceptance decision, and an explanation based on the trajectory. The generator changes the mechanic while keeping the persona, target dimension, and direction fixed. The new proposal goes through the same implementation checks and playtesting. Equation~\ref{eq:admission} compares it with the original game. If no suitable candidate exists or the feedback revision also fails, we retain the original game.

\begin{figure}[t]
\centering\includegraphics[width=0.85\linewidth]{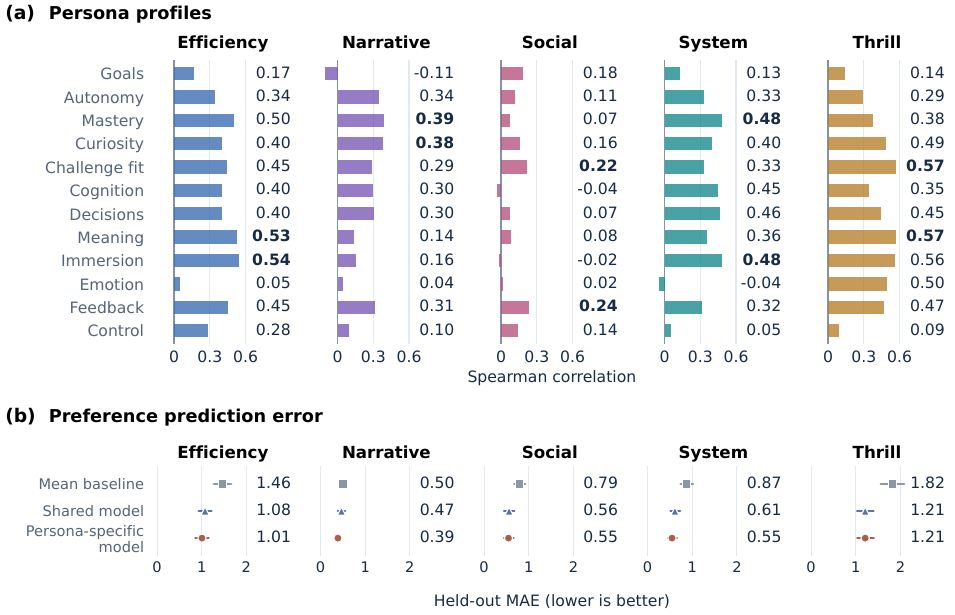}
\vspace{-5pt}
\caption{\textbf{Persona evaluation profiles and preference prediction.} (a) Experience dimensions relate differently to each persona's ratings. (b) Preference prediction from each persona's own experience scores, comparing a persona-mean baseline, shared dimension slopes with persona indicators, and separate persona models. Lower MAE indicates better prediction; intervals show uncertainty.}
\label{fig:persona-targets}
\vspace{-8pt}
\end{figure}

\vspace{-8pt}
\section{Experiments}
\label{sec:experiments}
\vspace{-5pt}
Our experiments connect three questions: what experience dimensions reveal about model evaluation, how persona-specific preference relationships define refinement targets, and whether the resulting game revisions receive higher post-play preference ratings. We examine each connection separately before following the complete process in a concrete case.
\vspace{-5pt}
\subsection{Experimental Setup}
\label{sec:benchmark-setup}
\vspace{-5pt}
\paragraph{Model evaluation.}
We benchmark GLM~\citep{glm5team2026glm5vibecodingagentic}, DeepSeek~\citep{deepseekai2026deepseekv41flashpushinglimitskv}, Kimi~\citep{kimiteam2026kimik2openagentic}, MiniMax~\citep{chen2026minimaxm2seriesminiactivations}, and Qwen~\citep{qwen3.8} on {91} card games with fixed rules and implementations. The models play against random opponents without persona prompts. Five neutral judges and five persona references assess the recorded trajectories with actor identities hidden. Overall ratings and 12 experience ratings are collected in separate calls. Appendix~\ref{app:benchmark} details model configurations and analysis sets.
\vspace{-8pt}
\paragraph{Persona prediction and refinement.}
We adopt five profiles from MeepleLM \citep{li2026meeplelm}: Efficiency Essentialist, Narrative Architect, Social Lubricator, System Purist, and Thrill Seeker. They provide reference evaluations for model comparison and serve as target players in game refinement. We validate preference predictions on held-out games and assess game revisions through persona playtesting and code implementation checks. Appendix~\ref{app:representation} describes the evaluation protocols.

\begin{table}[t]
\centering
\footnotesize
\setlength{\tabcolsep}{2.5pt}
\caption{\textbf{Agreement between adjacent game groups.} Group Gap is the mean rating difference, and Pair Agreement is the percentage of correctly ordered cross-group pairs.}
\label{tab:benchmark-consensus}
\begin{tabular}{lccccc}
\toprule
Judge Model & GLM-5.3 & DeepSeek-4.1flash & Kimi-K2.7 & MiniMax-M2.5 & Qwen3.8-Max\\
\midrule
Group Gap&1.87&1.10&1.69&1.61&1.59\\
Pair Agreement&96.0&78.4&88.0&91.5&93.7\\
\bottomrule
\vspace{-8pt}
\end{tabular}
\vspace{-8pt}
\end{table}
\begin{figure}[t]
\centering\includegraphics[width=.85\linewidth]{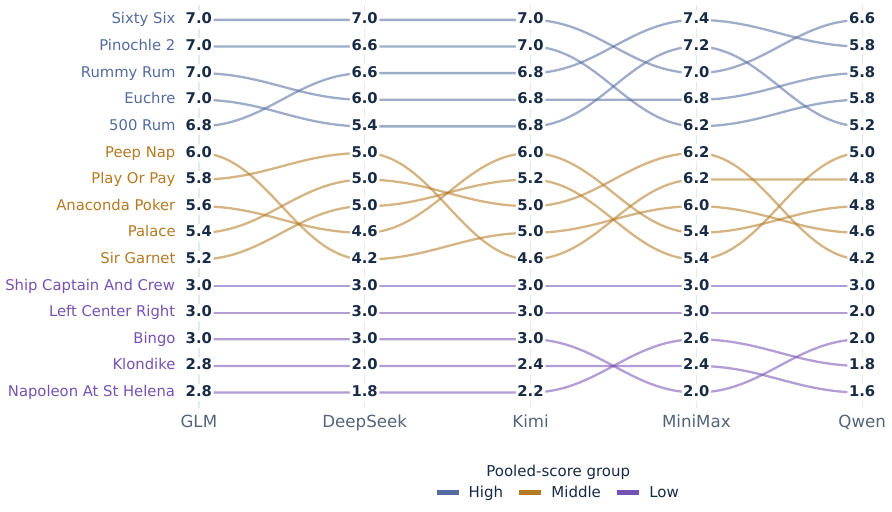}
\vspace{-8pt}
\caption{\textbf{Shared game preferences and local ranking differences.}}
\vspace{-8pt}
\label{fig:benchmark-ranking-examples}
\end{figure}

\vspace{-5pt}
\subsection{Model Evaluation through Experience Dimensions}
\label{sec:model-evaluation}
\vspace{-5pt}
We use experience dimensions to characterize how models evaluate recorded gameplay. Figure~\ref{fig:benchmark-experience} presents two complementary views: the associations between individual dimensions and overall ratings, and the similarity of these association profiles to persona references. 

We first examine whether models broadly agree on which games they rate highly. With complete rankings, four judges define high-, {middle-,} and low-rated groups for the fifth judge to compare. {Each group spans ten rank positions.} Each judge is held out in turn. All five favor the {higher}-rated group. {Averaging the high--middle and middle--low comparisons gives} mean gaps of {1.10-1.87} points and pairwise agreement of {78.4-96.0\%}, shown in Table \ref{tab:benchmark-consensus}. Figure~\ref{fig:benchmark-ranking-examples} illustrates this pattern with the highest, middle, and lowest five games by pooled score. Each column orders the displayed games by one judge's mean rating over actor sources, and lines connect the same game across judges. High-scoring games generally remain above low-scoring ones. At the level of overall ratings, the models show broad agreement.

We then compare how their overall ratings relate to experience dimensions and use persona references to interpret the resulting association profiles. Figure~\ref{fig:benchmark-experience}(a) shows positive mastery associations across models, alongside differences in immersion and challenge fit. The displayed mastery correlations range from {0.41} to {0.57}, placing mastery among the two largest correlations for every model. The other prominent associations differ: GLM has an immersion correlation of {0.56}, DeepSeek a challenge-fit correlation of {0.42}, and Qwen a meaning correlation of {0.54}. These patterns identify a common association with mastery while showing that the remaining dimensions relate differently to each model's overall ratings.

The persona references provide a way to interpret these differences in evaluation emphasis. Figure~\ref{fig:benchmark-experience}(b) expresses these evaluation patterns relative to the five persona references. \textit{System} has the largest {estimated} similarity share for GLM, DeepSeek, Kimi, and Qwen, while \textit{Thrill} has the largest share for MiniMax. Similarity describes a combination of evaluation emphases for each model. For example, MiniMax's shares are {25.9\%} for \textit{Thrill}, {22.0\%} for \textit{Narrative}, and {22.1\%} for \textit{System}.  These shares describe relative resemblance between evaluation profiles{, without implying positive raw correlations}. \textit{Social} {has the smallest normalized share for every neutral model}. DeepSeek yields {the largest share} for \textit{Social}, {13.2\%}, whereas the corresponding values for the other models {range from 7.7\% to 10.1\%}. This suggests that the pool of neutral judges may provide uneven coverage of different reference perspectives, highlighting the importance of persona-based perspectives.

\vspace{-5pt}
\subsection{Validating Experience-Based Preference Models}
\label{sec:persona-target-validation}
\vspace{-5pt}

The persona profiles in Figure \ref{fig:persona-targets}(a) characterize differences in evaluation emphasis on common benchmark inputs. For example, emotional challenge is positively associated with \textit{Thrill}'s ratings and negatively associated with \textit{System}'s. These patterns motivate preference predictors that combine experience dimensions for each persona. We next examine whether the same experience dimensions predict preference when games are played under the
target persona. This analysis uses a separate dataset of {455} persona-played
observations from {the same 91} games and five personas, with experience scores and
preference ratings collected separately.

\vspace{-8pt}

\paragraph{Preference prediction.}
We compare separate regularized linear predictors for each persona with a
persona-mean baseline and a shared predictor that uses common dimension
coefficients with persona-dependent score levels. Figure \ref{fig:persona-targets}(b) reports persona-wise prediction errors. Averaged equally over the five personas, the separate predictors achieve a mean absolute error (MAE) of {0.740}, compared with {1.090}
for the persona-mean baseline {and 0.785 for the shared predictor}. As shown in Figure~\ref{fig:persona-targets}(b), our {persona-specific} experience-based preference predictor {has lower MAE than the shared predictor} in {four of the five} personas{, with nearly identical errors for Thrill}. Appendix~\ref{app:preference-model} gives the procedures.

\vspace{-8pt}
\paragraph{\texorpdfstring{{Experience and persona ablations.}}{Experience and persona ablations.}}
Table~\ref{tab:persona-prediction} compares the experience sources {and persona conditioning} using identical observations {and held-out-game folds}. {MAE and $R^2$ are averaged over personas; Order measures accuracy for within-persona pairs with different observed ratings, with both games held out.} The nine PXI dimensions
achieve MAE {0.761} and ordering accuracy {81.8\%}, close to the combined
representation's {0.740} and {82.0\%}. Using only the three CORGIS challenge
dimensions yields MAE {0.905} and ordering accuracy {72.6\%}. {Removing both sources gives the persona-mean baseline. The - \textit{persona} ablation retains all 12 dimensions but shares both coefficients and intercept across personas, without persona indicators (MAE 0.880; Order 81.3\%).} PREFER uses {the relationships fitted on the original 49 games} to select
which dimensions to change and in which direction for each persona, guiding the game revisions evaluated in Section \ref{sec:refinement-results}.

\vspace{-5pt}
\subsection{Post-play Preference and Implementation Quality}
\label{sec:refinement-results}
\vspace{-5pt}
We compare \method{} with Cardiverse's game-generation workflow \citep{li2025cardiverse}, Reflexion's self-feedback refinement \citep{shinn2023reflexion}, ReCode's bounded recursive planning \citep{yu2025recode}, and PromptBreeder's mutation \citep{fernando2024promptbreeder}. We report {mean post-play preference} and {implementation metrics} in Cardiverse \citep{li2025cardiverse}: Success Rate, Execution Rate, code-rulebook embedding similarity (ECon), and play-rule consistency (PCon). {Appendices}~\ref{app:study-protocols} and \ref{app:implementation-protocol} {detail the adaptations, datasets, and protocols}.

\begin{table}[t]
\centering\small
\setlength{\tabcolsep}{2pt}
\caption{\textbf{Refinement and implementation outcomes.} Original games provide the preference reference. Bold marks the best method value, including ties.}
\label{tab:main}
\begin{tabular}{lccccc}
\toprule
Method & {Preference Rate} & Success Rate (\%) & Execution Rate (\%) & ECon & PCon\\
\midrule
Original games & {2.927} & -- & -- & -- & --\\
\midrule
Reflexion~\citep{shinn2023reflexion} & {2.787} & 83.33 & \textbf{99.96} & 0.656 & 6.514\\
PromptBreeder~\citep{fernando2024promptbreeder} & {2.740} & 73.33 & 99.86 & \textbf{0.670} & 5.407\\
Cardiverse~\citep{li2025cardiverse} & {2.820} & 76.67 & 99.52 & 0.658 & 5.991\\
ReCode~\citep{yu2025recode} & {3.053} & 83.33 & \textbf{99.96} & 0.659 & 6.648\\
\midrule
\method{} & \textbf{3.333} & \textbf{86.67} & \textbf{99.96} & 0.661 & \textbf{7.377}\\
\bottomrule
\end{tabular}
\end{table}
\begin{table}[t]
\centering
\begin{minipage}[t]{0.49\linewidth}
\vspace{0pt}
\centering\footnotesize
\setlength{\tabcolsep}{2pt}
\caption{\textbf{Experience {and persona ablations} in preference prediction.}}
\label{tab:persona-prediction}
\begin{tabular}{lccc}
\toprule
Predictor & MAE{$\downarrow$} & $R^2$ & \shortstack{Order(\%)}\\
\midrule
\method{} & 0.740 & 0.435 & 82.0\\
- \textit{CORGIS} & 0.761 & 0.426 & 81.8\\
{- \textit{persona}} & {0.880} & {0.030} & {81.3}\\
- \textit{PXI} & 0.905 & 0.222 & 72.6\\
- \textit{CORGIS\&PXI} & 1.090 & -0.022 & 50.0\\
\bottomrule
\end{tabular}
\end{minipage}\hfill
\begin{minipage}[t]{0.48\linewidth}
\vspace{0pt}
\centering\footnotesize
\setlength{\tabcolsep}{2.5pt}
\caption{\textbf{Mean paired preference change by persona.}}
\label{tab:persona}
\begin{tabular}{lcc}
\toprule
Persona & Directed refinement
        & With feedback\\
\midrule
Efficiency & $+0.333$ & $+0.400$\\
Narrative & $+0.067$ & $+0.067$\\
Social & $+0.233$ & $+0.267$\\
System & $+0.100$ & $+0.133$\\
Thrill & $+1.167$ & $+1.167$\\
\midrule
All personas & $+0.380$ & $+0.407$\\
\bottomrule
\end{tabular}
\end{minipage}
\vspace{-5pt}
\end{table}

As shown in Table \ref{tab:main}, \method{} achieves a mean preference of 3.333, a paired increase of 0.407 over the original games and {9.2\%} above the best baseline's mean of {3.053}. It also achieves the highest implementation success rate (86.67\%) and PCon (7.377), with execution rate tied at 99.96\%. The different method rankings under ECon and PCon highlight the value of checking actual gameplay alongside code-rulebook similarity.

Table~\ref{tab:persona} reports paired preference changes from the
original games {for both directed refinement and its feedback extension, averaged over 30 games per persona}.
Directed refinement achieves a mean gain of 0.380 points.
Persona-level mean gains range from 0.067 for \textit{Narrative} to
1.167 for \textit{Thrill}. These differences may reflect how readily persona preferences can be addressed through local rule edits. Such edits can reshape the immediate risks and consequences relevant to \textit{Thrill}, but may offer fewer opportunities to develop
thematic progression for \textit{Narrative} or improve strategic control and balance for \textit{System}. The feedback extension yields a small additional improvement,
increasing the mean paired preference gain from 0.380 to 0.407 points{, an additional 0.027 points}.
Overall, \method{} improves mean post-play preference in the evaluated
card games, with gains varying across personas.

\begin{figure}[t]
\centering\includegraphics[width=0.85\linewidth]{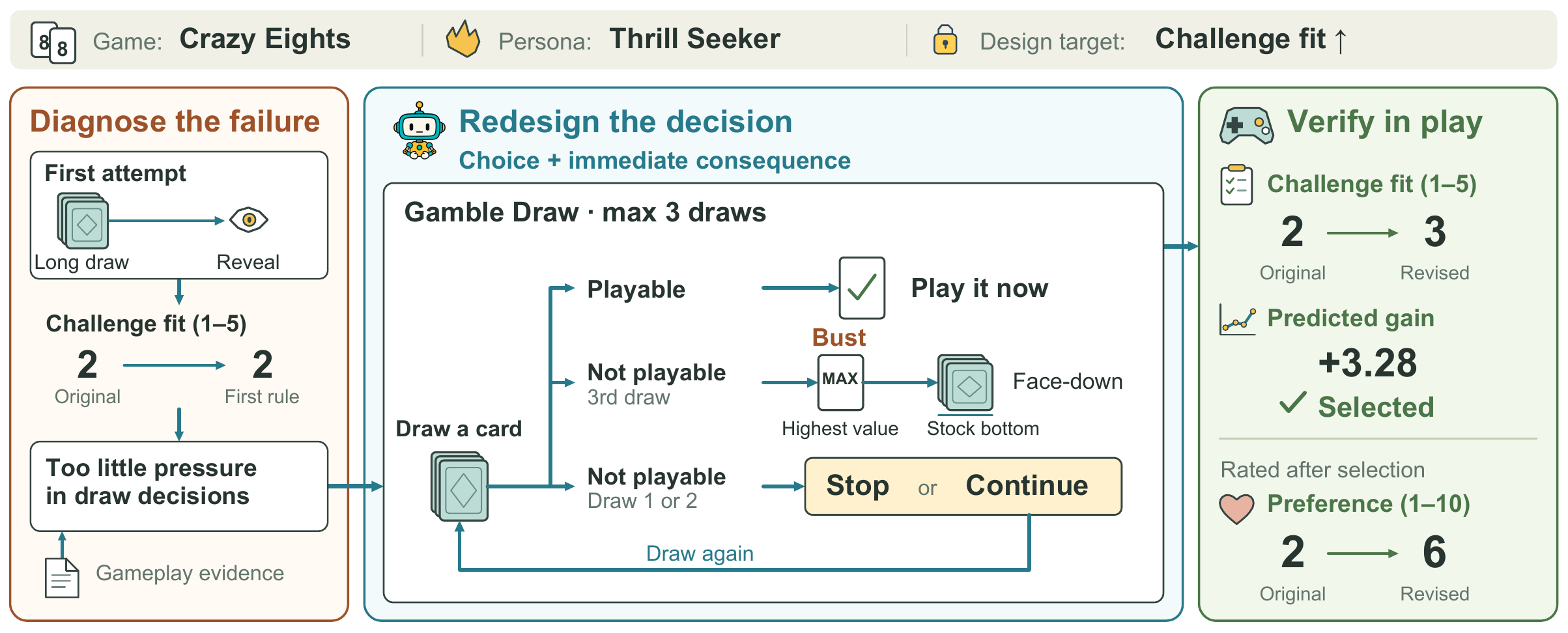}
\caption{\textbf{Feedback-guided revision in Crazy Eights.} The target persona and experience dimension remain fixed while playtesting feedback guides a new drawing mechanic.}
\label{fig:case}
\vspace{-5pt}
\end{figure}

\vspace{-5pt}
\subsection{Case Study: From an Experience Target to a Rule Change}
\label{sec:case}
\vspace{-5pt}

Figure~\ref{fig:case} follows a Crazy Eights revision for Thrill Seeker with a fixed challenge-fit target. Revealing a player's hand after extended drawing leaves challenge fit unchanged. Feedback then guides a stop-or-continue drawing mechanic with a three-draw limit. In the resulting playtest, challenge fit rises from 2 to 3 and post-selection preference from 2 to 6. The case illustrates how feedback guides a concrete mechanic change toward the same experience target. Appendix~\ref{app:case-evidence} gives the full rule changes and evaluation record.

\section{Conclusion}
\method{} uses shared experience dimensions to connect model evaluation with persona-guided game refinement. The dimensions reveal differences hidden by overall ratings, predict persona preference on held-out games, and define concrete modification targets. Delivered revisions receive higher mean post-play preference than the original games and evaluated baselines. The implementation evaluation also reports higher success and play-rule consistency. Together, these results demonstrate how experience measurements can inform concrete, executable game revisions.

\clearpage

\subsection*{AI use statement}
\label{sec:ai-use}
Generative AI tools assisted manuscript polishing, game code generation, and icon creation. The authors reviewed the manuscript content, checked the generated game code for executability, and reviewed the use of the generated icons. The use of LLMs as game-playing agents, experience evaluators, persona-conditioned judges, and rule and code generators within the research is described in Sections~\ref{sec:method} and~\ref{sec:experiments}. The authors take responsibility for the final content of this work, including AI-assisted text, code, and visual assets.

\bibliography{references}
\bibliographystyle{iclr2027_conference}
\clearpage
\appendix
\raggedbottom
\renewcommand{\floatpagefraction}{0.85}
\setcounter{topnumber}{4}\setcounter{totalnumber}{6}
\setlength{\textfloatsep}{12pt plus 2pt minus 2pt}
\makeatletter\setlength{\@fptop}{0pt}\setlength{\@fpsep}{16pt}\setlength{\@fpbot}{0pt plus 1fil}\makeatother
\section{Benchmark Details}
\label{app:benchmark}

\subsection{Data, Evaluators, and Analysis Sets}
The neutral benchmark contains {91} fixed games and five actor configurations. We match requests by the exact rule and focal-player evidence texts, deduplicate identical inputs, and require every instrument needed for a comparison. The common set has {363} unique inputs from {91} games with overall and 12-dimensional ratings from all five neutral judges and all five persona references. Overall replies {with insufficient evidence} remain missing; they are not replaced by zero. Persona scoring includes the completed supplementary questionnaires. {Completed play includes trajectories stopped at the decision or time limit.}

The focal-player evidence records visible states, legal actions, selected actions, and outcomes. Initial play fixes seat 0 and seed 0, with random agents at other seats, limits of 500 focal decisions and four hours, and at most one model-selected recovery from an invalid action. Evaluation requests hide actor identity and collect dimensions and overall scores separately. The overall prompt concerns the recorded experience, excluding replay intention and recommendation.

The actor configurations are GLM-5.3, DeepSeek-V4.1-Flash, Kimi-K2.7-Code, MiniMax-M2.5, and Qwen3.8-Max. The Qwen judge is Qwen3.8-Max-0902. The original external-rating batch requests temperature 0 for Kimi, DeepSeek, MiniMax and Qwen. Reasoning settings and output limits differ across services.

\subsection{Dimension Associations and Within-game Checks}
\label{app:benchmark-within}
For each evaluator, we correlate its dimension ratings with its separately obtained overall scores, using Spearman correlation with average ranks for ties. Table~\ref{tab:benchmark-ring-values} gives the values underlying the model rings and persona profiles in the main text. These pooled associations can reflect differences between games. To inspect variation within games, we rank scores and dimensions across the common set, subtract the corresponding game means from their ranks, and correlate the centered ranks. Table~\ref{tab:benchmark-within} reports these within-game associations. Intervals use 2,000 whole-game bootstrap samples. {Simultaneous intervals account for multiple model-pair and dimension comparisons.}

\begin{table}[!htbp]
\centering\scriptsize
\setlength{\tabcolsep}{2.6pt}
\caption{\textbf{Associations between experience dimensions and overall ratings.} Spearman correlations characterize evaluation emphasis across models and personas. E, N, S, Y, and T denote Efficiency, Narrative, Social, System, and Thrill.}
\label{tab:benchmark-ring-values}
\begin{tabular}{lrrrrrrrrrr}
\toprule
Dimension & GLM & DS & Kimi & MM & Qwen & E & N & S & Y & T\\
\midrule
Goals/rules & 0.12 & 0.06 & -0.01 & 0.03 & 0.12 & 0.17 & -0.11 & 0.18 & 0.13 & 0.14\\
Autonomy & 0.44 & 0.20 & 0.17 & 0.24 & 0.35 & 0.34 & 0.34 & 0.11 & 0.33 & 0.29\\
Mastery & 0.57 & 0.41 & 0.48 & 0.43 & 0.56 & 0.50 & 0.39 & 0.07 & 0.48 & 0.38\\
Curiosity & 0.49 & 0.18 & 0.27 & 0.33 & 0.42 & 0.40 & 0.38 & 0.16 & 0.40 & 0.49\\
Challenge fit & 0.06 & 0.42 & 0.00 & 0.38 & 0.20 & 0.45 & 0.29 & 0.22 & 0.33 & 0.57\\
Cognitive challenge & 0.45 & 0.32 & 0.26 & 0.31 & 0.49 & 0.40 & 0.30 & -0.04 & 0.45 & 0.35\\
Decision challenge & 0.42 & 0.32 & 0.18 & 0.31 & 0.41 & 0.40 & 0.30 & 0.07 & 0.46 & 0.45\\
Meaning & 0.47 & 0.28 & 0.15 & 0.38 & 0.54 & 0.53 & 0.14 & 0.08 & 0.36 & 0.57\\
Immersion & 0.56 & 0.15 & 0.21 & 0.37 & 0.49 & 0.54 & 0.16 & -0.02 & 0.48 & 0.56\\
Emotional challenge & 0.03 & 0.01 & -0.02 & 0.34 & 0.11 & 0.05 & 0.04 & 0.02 & -0.04 & 0.50\\
Progress feedback & 0.51 & 0.19 & 0.20 & 0.22 & 0.32 & 0.45 & 0.31 & 0.24 & 0.32 & 0.47\\
Ease of control & 0.15 & 0.03 & -0.03 & 0.09 & 0.08 & 0.28 & 0.10 & 0.14 & 0.05 & 0.09\\
\bottomrule
\end{tabular}
\end{table}

\begin{table}[!htbp]
\centering\small
\setlength{\tabcolsep}{4pt}
\caption{\textbf{Experience associations within games.} Correlations of game-centered ranks describe how experience ratings and overall judgments vary among trajectories of the same game.}
\label{tab:benchmark-within}
\begin{tabular}{lrrrrr}
\toprule
Dimension & GLM & DeepSeek & Kimi & MiniMax & Qwen\\
\midrule
Goals/rules & 0.174 & -0.001 & 0.114 & 0.031 & 0.139\\
Autonomy & 0.132 & 0.150 & 0.025 & 0.117 & 0.144\\
Mastery & 0.321 & 0.233 & 0.245 & 0.210 & 0.270\\
Curiosity & 0.236 & 0.042 & 0.121 & 0.122 & 0.062\\
Challenge fit & 0.080 & 0.251 & -0.005 & 0.134 & -0.036\\
Cognitive challenge & 0.210 & 0.112 & 0.019 & 0.090 & 0.152\\
Decision challenge & 0.136 & 0.175 & 0.004 & 0.104 & 0.046\\
Meaning & 0.259 & 0.135 & 0.059 & 0.102 & 0.211\\
Immersion & 0.261 & 0.090 & 0.035 & 0.095 & 0.140\\
Emotional challenge & -0.012 & 0.015 & 0.012 & 0.134 & -0.001\\
Progress feedback & 0.359 & 0.199 & 0.118 & 0.090 & 0.124\\
Ease of control & 0.180 & 0.144 & 0.059 & 0.067 & 0.158\\
\bottomrule
\end{tabular}
\end{table}

\paragraph{Model and persona profiles.}
Across the shared inputs, mastery is positively associated with overall ratings for every model ($\rho={0.413}$-${0.574}$). Immersion has correlation {0.560} for GLM, compared with {0.148} for DeepSeek and {0.212} for Kimi. Challenge fit has correlations {0.419} for DeepSeek and {0.380} for MiniMax, compared with {$-0.003$} for Kimi. For the Thrill persona, challenge fit, meaning, immersion, and emotional challenge correlate with overall ratings at {0.567, 0.571, 0.562, and 0.496}, respectively; System's emotional-challenge correlation is {$-0.044$}. These values describe pooled associations across the recorded games and trajectories.

\begin{figure}[!htbp]
\centering\includegraphics[width=\linewidth]{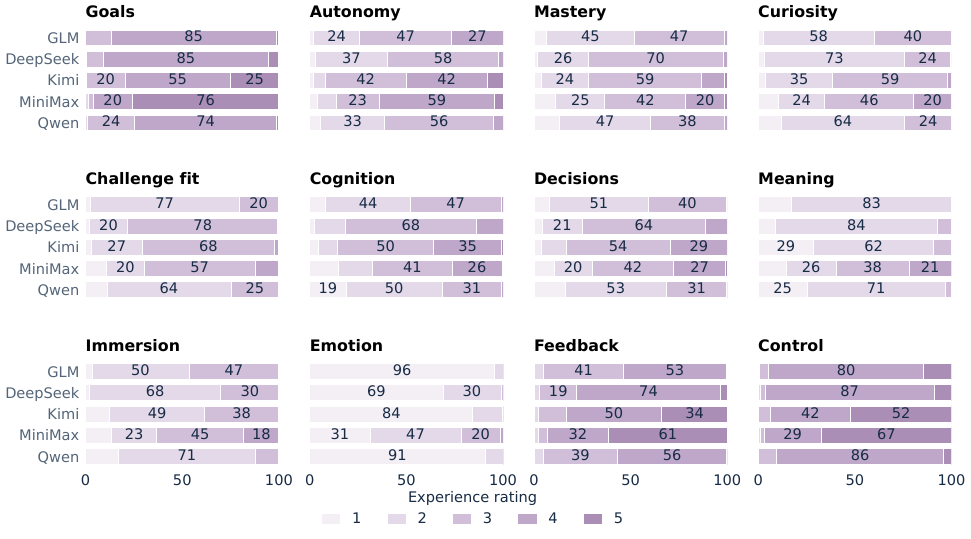}
\caption{\textbf{How evaluators use the experience rating scale.} Rating distributions reveal differences in scoring levels across models, personas, and experience dimensions. Colors identify the five rating levels.}
\label{fig:rating-distributions}
\end{figure}
\FloatBarrier
\subsection{Scoring Level and Same-input Judgments}
\label{app:benchmark-case}
\label{app:benchmark-judgments}
We compare all ten judge pairs on each common input, yielding {3,630} comparisons. Dimension distance is the mean absolute difference across the 12 ratings. We estimate each pair's usual signed overall-score difference from all other games, weighting games equally, subtract this offset from the held-out comparison, and then take the absolute adjusted gap. Bootstrap samples refit the offset while excluding every copy of the held-out game.
{Applying thresholds of} dimension MAE at most 0.5 and overall gap at least three {identifies} {63 pairs from 43 inputs and 30 games}. For the {1,834} pairs with dimension MAE at most 0.5, the share with gap at least two falls from {31.5\%} to {8.8\%} after adjustment. The reduction is {22.6} percentage points. For a gap of at least three, the corresponding shares are {3.4\% and 1.4\%}.

\FloatBarrier
\begin{figure}[!htbp]
\centering\includegraphics[width=\linewidth]{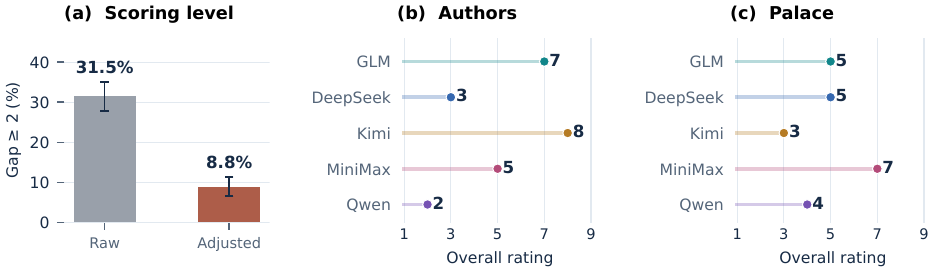}
\caption{\textbf{Scoring levels and judgments of the same play.} (a) Accounting for habitual scoring differences reduces large overall-score gaps among judges with similar dimension ratings. (b,c) Authors and Palace illustrate differing evaluations of identical gameplay.}
\label{fig:benchmark-judgments}
\end{figure}
In Authors, 26 successful requests produce a 13-0 win. GLM assigns 7 for smooth progress, whereas Qwen assigns 2 and emphasizes limited challenge and uncertainty. In Palace, Kimi assigns 3 and emphasizes the loss and slow hand reduction; MiniMax assigns 7 and values varied, coherent interaction. In Cold Hands Poker, one forced ante ends at a focal payoff of $-5$: DeepSeek assigns 5 for an empty but not actively frustrating experience, whereas Qwen assigns 2 for a lack of meaningful play. In Arlington, GLM assigns 8 and values agency and progress, whereas DeepSeek assigns 5 because procedural evidence offers limited support for strong subjective engagement.
\subsection{Persona Reference Similarity}
\label{app:benchmark-personas}
Let $v_r$ contain evaluator $r$'s 12 dimension-overall correlations. We center each vector across dimensions and compare it with each persona profile:
\begin{equation}
\begin{split}
c_r&=v_r-\overline v_r\mathbf{1},\\
s_{m,p}&=\frac{c_m^\top c_p}{\lVert c_m\rVert\lVert c_p\rVert},\qquad
q_{m,p}={\frac{1+s_{m,p}}{\sum_{p'}(1+s_{m,p'})}}.
\end{split}
\label{eq:benchmark-similarity}
\end{equation}
The raw similarity is Pearson correlation between association profiles and may be negative. {We map similarity from $[-1,1]$ to $[0,1]$ using $(1+s)/2$, then normalize across the five personas to obtain $q$. Negative similarities are retained by this transformation. The shares describe relative resemblance among the five reference profiles; equal similarities yield a 20\% share for each persona.} Constant profiles and {a zero denominator} remain undefined. Whole-game bootstrap samples recompute the associations and similarities; 2,000 samples provide the {intervals shown in the figures}.

\begin{figure}[!htbp]
\centering\includegraphics[width=\linewidth]{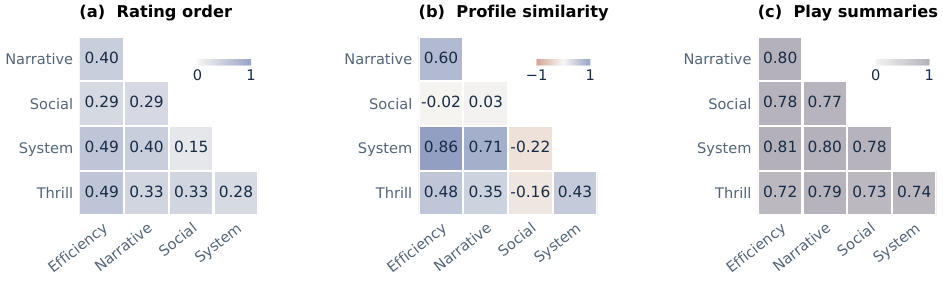}
\caption{\textbf{Relations among the five personas.} The panels compare agreement in overall ratings, resemblance of dimension-association profiles, and proximity of gameplay behavior. These complementary views describe different aspects of persona similarity.}
\label{fig:persona-internal}
\end{figure}

Figure~\ref{fig:persona-internal} also compares gameplay behavior. It averages within-game rank proximity among the ten player configurations across {all 91 games}. {Ties receive average ranks, including when all configurations share the same descriptor value.}

\begin{table}[!htbp]
\centering\small
\setlength{\tabcolsep}{4pt}
\caption{\textbf{Relative resemblance to persona evaluation profiles.} Normalized similarity shares (\%) summarize the relative emphasis of each persona within a model's evaluation profile.}
\label{tab:benchmark-similarity}
\begin{tabular}{lrrrrr}
\toprule
Judge & Efficiency & Narrative & Social & System & Thrill\\
\midrule
GLM & 24.3 & 22.4 & 10.1 & 25.3 & 17.9\\
DeepSeek & 22.5 & 22.3 & 13.2 & 23.3 & 18.7\\
Kimi & 23.0 & 24.2 & 9.9 & 25.2 & 17.7\\
MiniMax & 21.7 & 22.0 & 8.3 & 22.1 & 25.9\\
Qwen & 24.5 & 21.8 & 7.7 & 25.8 & 20.2\\
\bottomrule
\end{tabular}
\end{table}

\begin{figure}[!htbp]
\centering\includegraphics[width=\linewidth]{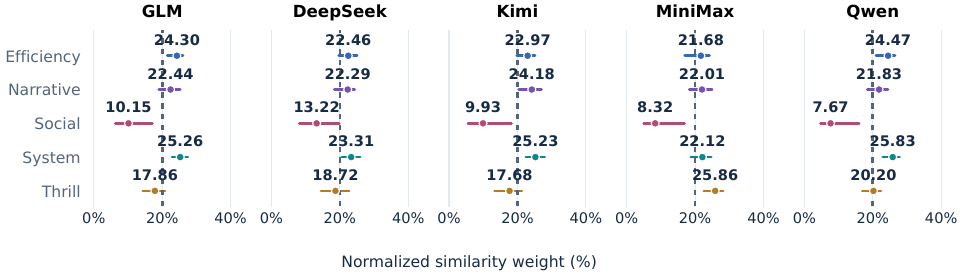}
\caption{\textbf{Relative persona resemblance and its uncertainty.} Normalized similarity shares show each model's relative resemblance to the five reference profiles. The 20\% line indicates equal shares; intervals show uncertainty.}
\label{fig:persona-raw-similarity}
\end{figure}

\FloatBarrier
\subsection{Persona Scores Reconstructing Neutral Ratings}
\label{app:benchmark-mixture}
This supplementary analysis predicts a neutral judge's overall score from the five persona scores on the same input. It is distinct from predicting a target persona's preference from experience dimensions in Section~\ref{sec:persona-target-validation}. The reconstruction uses {363} inputs from {91} games. Within each training fold, persona scores and the target are standardized with {means and standard deviations that weight games equally}. Nonnegative mixture weights sum to one:

\begin{equation}
\min_{\beta_j}\sum_i a_i\left(t_{i,j}-\sum_p\beta_{j,p}z_{i,p}\right)^2,
\qquad \beta_{j,p}\geq0,\quad\sum_p\beta_{j,p}=1,
\label{eq:benchmark-mixture}
\end{equation}
where $a_i$ is {the reciprocal of} the number of inputs in its game. Each game is held out in turn, and predictions are returned to the judge's original scale. Baselines use the training mean, equal persona weights, the single persona selected by training error, and a fitted Efficiency-Thrill mixture. MSE is averaged first within game and then over games. Error-contrast intervals resample the {91} fixed held-out game losses 2,000 times; coefficient intervals use 1,000 whole-game refits.

\begin{figure}[!htbp]
\centering
\begin{minipage}[t]{0.49\linewidth}
\vspace{0pt}
\centering\footnotesize
\setlength{\tabcolsep}{2.5pt}
\captionof{table}{\textbf{Reconstructing neutral ratings from persona scores.} Held-out MSE compares alternative combinations of persona evaluations. Lower values indicate better prediction.}
\label{tab:benchmark-mixture}
\begin{tabular}{lrrrrr}
\toprule
Judge & Mean & Equal & Single & \shortstack{Five\\personas} & \shortstack{Two\\personas}\\
\midrule
GLM & 2.292 & 1.116 & 1.602 & \textbf{1.112} & 1.306\\
DeepSeek & 1.507 & \textbf{1.192} & 1.592 & 1.227 & 1.398\\
Kimi & 2.560 & \textbf{1.668} & 2.393 & 1.684 & 1.982\\
MiniMax & 2.412 & \textbf{1.534} & 2.305 & 1.555 & 1.951\\
Qwen & 1.606 & 0.603 & 0.845 & \textbf{0.552} & 0.735\\
\bottomrule
\end{tabular}
\end{minipage}\hfill
\begin{minipage}[t]{0.48\linewidth}
\vspace{0pt}
\centering\includegraphics[width=\linewidth]{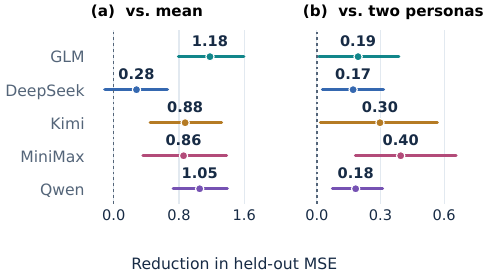}
\caption{\textbf{Predictive value of persona score combinations.} Positive error reductions favor the five-persona fit over (a) the mean baseline and (b) the Efficiency--Thrill fit. Intervals show uncertainty.}
\label{fig:benchmark-prediction}
\end{minipage}
\end{figure}

\begin{figure}[!htbp]
\centering\includegraphics[width=\linewidth]{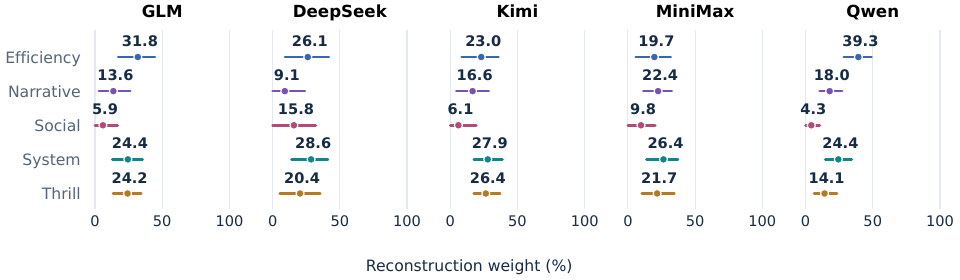}
\caption{\textbf{Persona contributions to neutral-score reconstruction.} Fitted weights describe how persona ratings combine to predict each neutral judge's scores. Intervals show uncertainty in these relative contributions.}
\label{fig:persona-fitted-weights}
\end{figure}
\FloatBarrier
\subsection{Own-source Rating Contrasts}
\label{app:benchmark-own}
\begin{figure}[!htbp]
\centering
\begin{minipage}[t]{0.48\linewidth}
\vspace{0pt}
\centering\footnotesize
\setlength{\tabcolsep}{2pt}
\captionof{table}{\textbf{Ratings of a model's own gameplay relative to other actors.} Adjusted contrasts account for actor preferences shared by the judge panel.}
\label{tab:benchmark-own}
\begin{tabular}{lrr}
\toprule
Judge & \shortstack{Raw\\mean} & \shortstack{Adjusted\\mean}\\
\midrule
GLM & +0.159 & +0.048\\
DeepSeek & -0.206 & +0.026\\
Kimi & +0.113 & +0.059\\
MiniMax & +0.080 & +0.147\\
Qwen & +0.080 & -0.056\\
\bottomrule
\end{tabular}
\end{minipage}\hfill
\begin{minipage}[t]{0.49\linewidth}
\vspace{0pt}
\centering\includegraphics[width=\linewidth]{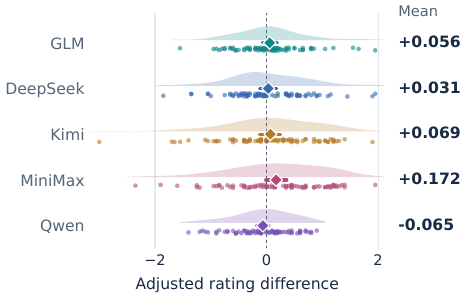}
\caption{\textbf{Own-source rating differences across games.} Adjusted contrasts vary across games, while average effects remain compatible with zero. Diamonds indicate means and lines show uncertainty intervals.}
\label{fig:own-source-distribution}
\end{minipage}
\end{figure}

For each game and judge, we subtract its average rating of other actors from its rating of its own actor's trajectory. We then {subtract the panel's mean contrast for that actor}. This adjustment accounts for actor outputs that all judges prefer; actor identities remain hidden during evaluation. All adjusted intervals {shown in Figure~\ref{fig:own-source-distribution}} include zero. The Qwen judge and actor have different service configurations, so ``own source'' denotes their paired family label, not an identical endpoint.

\FloatBarrier
\subsection{Game Rankings and Broad Consensus}
\label{app:benchmark-rankings}
\begin{figure}[!htbp]
\centering\includegraphics[width=\linewidth]{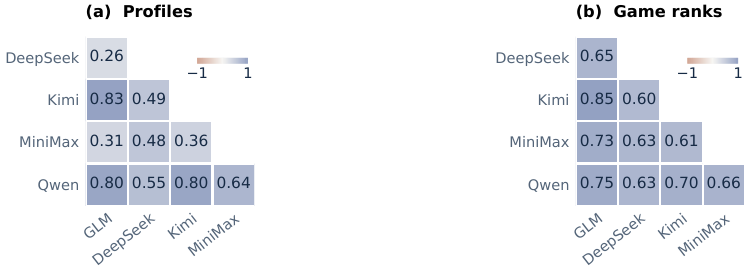}
\caption{\textbf{Similarity in evaluation emphasis and game rankings.} Models can agree on which games score highly while differing in how experience dimensions relate to their overall judgments.}
\label{fig:benchmark-two-layer}
\end{figure}

Game scores average the five actor sources equally within each of {91} complete games. Figure~\ref{fig:benchmark-two-layer} compares rank similarity at two levels on matched data. The left panel correlates the ordering of the 12 dimension associations; the right correlates game scores. Figure~\ref{fig:benchmark-ranking-examples} displays 15 games selected from the high, middle, and low pooled-score groups. Pairwise game-ranking correlations range from {0.599 to 0.847}. For GLM and DeepSeek, the game-ranking correlation is {0.647} and the dimension-profile rank correlation is {0.259} on the matched set.

{For Table~\ref{tab:benchmark-consensus}, the other four judges define high, middle, and low groups, each spanning ten rank positions. Fractional weights preserve ties at group boundaries. Group Gap and Pair Agreement average the high--middle and middle--low comparisons; tied scores receive half credit in pair agreement.}

\FloatBarrier
\subsection{Evaluation Resemblance and Produced Play}
\label{app:benchmark-alignment}
{Evaluation resemblance and output endorsement} are matched on {all 91 benchmark games, covering 363 unique inputs}. For a model-persona pair, evaluation resemblance is the Spearman correlation of their overall ratings, centered over personas within the model. Output endorsement is the persona's rating of that model's play after removing game-specific model and persona mean effects. This resemblance measure uses overall ratings, rather than the dimension-profile shares in Figure~\ref{fig:benchmark-experience}. The joint game-bootstrap analysis recomputes both directions. The correlation across the 25 model-persona cells is {$0.046$}.

\begin{figure}[!htbp]
\centering\includegraphics[width=\linewidth]{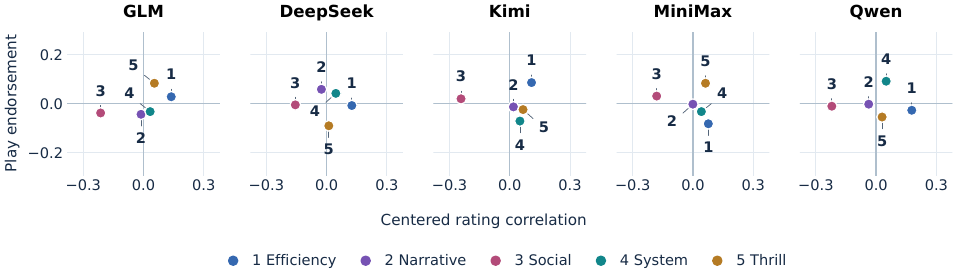}
\caption{\textbf{Evaluation resemblance and endorsement of model-produced play.} Similarity in rating patterns shows no clear overall association with relative endorsement of a model's gameplay. Colors and numbers identify personas.}
\label{fig:persona-alignment}
\end{figure}
\begin{figure}[!htbp]
\centering\includegraphics[width=\linewidth]{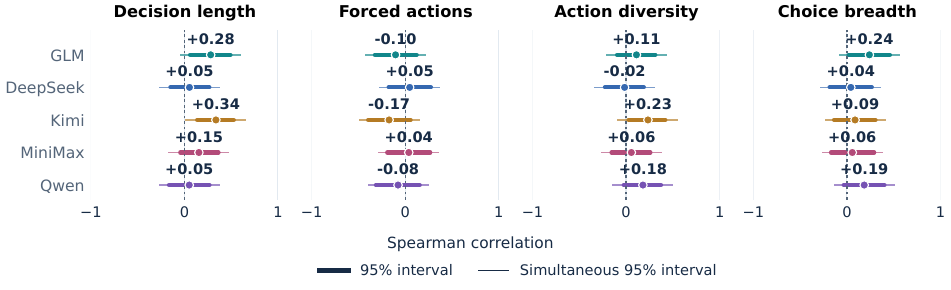}
\caption{\textbf{Observed gameplay characteristics and overall ratings.} Associations summarize how decision length and action-choice characteristics relate to scores. Thick and thin lines show individual and simultaneous uncertainty intervals.}
\label{fig:observed-play-features}
\end{figure}

A separate exploratory analysis relates log-derived play descriptors to game ratings {on all 91 benchmark games}. We average the five actor sources equally per game, then correlate game-average scores with {log-transformed decision count, forced-action share, action-type diversity, and the mean log number of choices at non-forced decisions}. {For trajectories with no free decisions, free-action diversity and choice breadth are set to zero.} {Figure~\ref{fig:observed-play-features}} reports individual and simultaneous uncertainty.

\FloatBarrier

\section{Experience Representation and Evaluation Details}
\label{app:representation}

\paragraph{Evaluation datasets.}
\label{app:study-protocols}
Preference-model validation uses {91} games played under each of the five personas, giving {455} observations. The refinement records cover 30 games and five personas (150 conditions); the preference comparison in Table~\ref{tab:main} uses {the same 150 conditions, retaining the original-game score when a revision fails}. Its implementation columns come from a separate set of runs on 30 shared-rule tasks. The preference runs evaluate games delivered to the target persona, while the implementation runs compare code generated for fixed target rulebooks. {Reflexion and ReCode means use the reported aggregate scores for the evaluated conditions and the original-game scores for failed revisions.} Appendix~\ref{app:historical-refinement} describes the {paired preference outcomes}, and Appendix~\ref{app:implementation-protocol} details the implementation protocol. {Preference predictors used to guide these historical revisions remain fitted on the original 49-game dataset; the expanded validation set does not change the recorded revisions.}

Table~\ref{tab:dimensions} groups the selected constructs by their source instrument. PXI contributes functional and psychosocial properties \citep{vanden2020development}, and CORGIS contributes three forms of challenge \citep{denisova2020measuring}. We adapt these concepts to trajectory-grounded scores on a 1-5 scale, allowing the same preference-model family to operate across game-specific rules and action spaces.

\begin{table}[ht]
\centering\small
\caption{\textbf{Experience constructs selected from two instruments.} PXI supplies functional and psychosocial properties; CORGIS distinguishes sources of challenge. The selected constructs are adapted to the persona's played trajectory.}
\label{tab:dimensions}
\vspace{\baselineskip}
\begin{tabularx}{\linewidth}{lX}
\toprule
Selected construct & Trajectory-based interpretation\\
\midrule
\multicolumn{2}{l}{\textbf{Player Experience Inventory (PXI)}}\\
Goals and rules & Clarity of objectives and the rules needed to pursue them\\
Challenge fit & Fit between the challenge and the intended player's experience\\
Progress feedback & Perceptible information about progress and consequences\\
Ease of control & Ease of expressing the intended game action\\
Autonomy & Meaningful opportunities to choose how to act\\
Mastery & Opportunities to apply and develop competence\\
Curiosity & Interest in discovering game states, outcomes, or possibilities\\
Meaning & Perceived purpose and significance of play\\
Immersion & Absorption in the unfolding game\\
\midrule
\multicolumn{2}{l}{\textbf{Challenge Originating from Recent Gameplay Interaction Scale (CORGIS)}}\\
Cognitive challenge & Demands on reasoning and understanding\\
Decision challenge & Demands created by consequential alternatives\\
Emotional challenge & Emotional stakes and demands during play\\
\bottomrule
\end{tabularx}
\end{table}

\paragraph{Preference ratings.}
The post-play evaluator receives the persona profile, rule context, and concrete events from the played trajectory. The anchors are: 1, actively avoid another play; 3, unlikely to replay; 5, neutral or mixed; 7, willing to replay; 9, strongly prefers replaying or choosing this game; and 10, exceptional fit. A rating includes trajectory-specific evidence. The median of three temperature-zero evaluations of each trajectory is the reported preference score. The five personas are LLM player agents.

\paragraph{Experience measurement during refinement.}
The measurement prompt receives the persona profile, rulebook, focal player, and played trajectory and returns a 1-5 score with trajectory evidence for each dimension. The selected attributes are measured in two fixed prompt batches. Three temperature-zero measurements are combined by the median for each dimension. Long play records are divided at turn boundaries, with observation-count-weighted scores across the selected chunks within each call.

\paragraph{Refinement protocol.}
Each condition starts from its frozen original game and preference score. A candidate's identity comprises game, persona, target dimension, and direction. Unchanged historical candidates reuse their corresponding evaluated outcome; newly selected candidates receive post-selection evaluation. The complete result includes every condition, assigning the original score when the procedure retains the original game. {The same fallback applies to failed baseline revisions, which contribute zero preference change in the 30-game comparison.} One gameplay trajectory per selected condition supplies the three preference ratings.

\subsection{Preference Predictor}
\label{app:preference-model}

For persona $p$, the preference predictor takes the form
\begin{equation}
f_p(x)=b_p+w_p^\top x,
\label{eq:preference}
\end{equation}
where $b_p$ is the intercept and $w_p$ contains the dimension-specific coefficients. The model is fitted with regularization on paired experience and preference measurements from other games before being used to guide refinement of the current game. We express the coefficients in the original scoring units, so that $w_{pj}$ represents the predicted preference change associated with a unit increase in dimension $j$, holding the remaining dimensions fixed.

\paragraph{Data and validation.}
The {validation} dataset contains {455} complete experience-preference observations: one persona-played record for every combination of {91} games and five personas. A separate ridge model is fitted for each persona with regularization coefficient 10. Standardization and median imputation use training-fold statistics only; the current data have complete dimensions. Absolute-score prediction holds out the evaluated game across all personas. Pairwise validation excludes both compared games and evaluates within-persona pairs with distinct observed preference labels. Both predictions in a pair come from the same fitted model; a tied prediction receives half credit. Regression metrics average the five persona-specific results equally. Ordering accuracy pools {14,785} unequal-label pairs.

\paragraph{Shared relationship baseline.}
The shared model is a pooled ridge regression with one common coefficient per dimension and five persona one-hot indicators. All feature columns, including the indicators, are standardized within each training fold; all coefficients use the same ridge penalty 10. Thus, it permits persona-dependent score levels while sharing the dimension slopes. The mean baseline predicts each persona's training mean. {The - \textit{persona} ablation uses one pooled ridge model with the same 12 experience dimensions but no persona indicators, so both slopes and intercept are shared.} Every model excludes the same held-out game or game pair. {Figure~\ref{fig:persona-targets}(b) compares separate models with the shared and mean baselines; Table~\ref{tab:persona-prediction} reports the ablations.} Uncertainty {shown in the figures} for absolute errors resamples whole games from the fixed held-out predictions (2,000 replicates). The pooled comparator and separate models use the same fixed penalty but differ in training sample size and feature scaling.

\paragraph{Prediction results.}
Separate persona models achieve MAE {0.740} versus {1.090} for the persona-mean baseline, a paired error reduction of {0.350} points. The shared model achieves MAE {0.785}; the paired error reduction from shared to separate models is {0.046} points. Mean persona-wise $R^2$ is {0.435} for the separate models, and their ordering accuracy is {82.0\%}, compared with {81.6\%} for the shared model. Both games in each ordering comparison are excluded from fitting. {Without persona indicators, MAE is 0.880, mean persona-wise $R^2$ is 0.030, and ordering accuracy is 81.3\%.}

\paragraph{Experience-source comparison.}
PXI uses goals/rules, challenge fit, progress feedback, ease of control, autonomy, mastery, curiosity, meaning, and immersion; CORGIS uses cognitive, decision, and emotional challenge. All source comparisons retain identical observations, labels, fitting procedures, and validation folds. Adding CORGIS to PXI {reduces MAE by 0.021 points}.

\begin{figure}[!htbp]
\centering\includegraphics[width=0.60\linewidth]{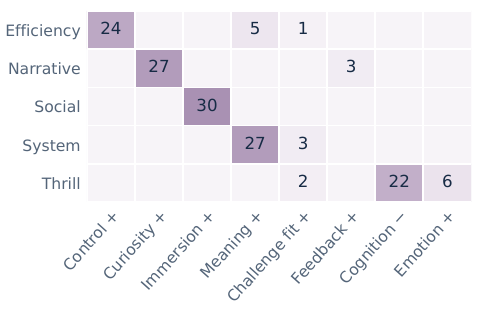}
\caption{\textbf{Preferred refinement directions across personas.} First-ranked feasible targets reveal different priorities in the refinement experiment. Plus and minus signs denote intended increases and decreases in an experience dimension.}
\label{fig:persona-target-counts}
\end{figure}

\paragraph{Target frequencies.}
Figure~\ref{fig:persona-target-counts} counts the first-ranked feasible target over 30 games per persona {in the historical refinement experiment}. Efficiency first targets higher ease of control in 24 games, Narrative higher curiosity in 27, Social higher immersion in all 30, and System higher meaning in 27. Thrill first targets lower cognitive challenge in 22 games. The ranking uses the conditional coefficients of $f_p$ in the original scoring units after filtering for feasible changes.

\paragraph{Verification of actual refinement targets.}
The original refinement records {fix} three feasible targets per game-persona condition, ranked by the magnitude of the original-unit coefficient, with direction given by its sign. We independently refit the leave-current-game models on the {original 49-game dataset (245 observations)}, divide each standardized coefficient by its training feature scale, and match the saved coefficients for all 150 conditions and all 450 ranked targets to numerical tolerance $10^{-8}$. Figure~\ref{fig:persona-target-counts} counts the first-ranked feasible target, not the eventually accepted target. In Crazy Eights, for example, Efficiency first targets meaning ($w=0.583$), Narrative curiosity ($0.332$), Social immersion ($0.843$), System meaning ($0.422$), and Thrill lower cognitive challenge ($-0.721$). Thrill's challenge-fit target is third in that game's frozen search order ($w=0.643$). The target is not changed using an observed post-intervention preference score.

\section{Supplementary Refinement Results}
\label{app:accounting}
\label{app:historical-refinement}

\paragraph{Paired outcomes {across 30 games}.}
Each {of the 30 games} contributes all five personas. {With the feedback extension,} PREFER improves {39} of the {150} original-game conditions, ties {107}, and worsens four.

Directed refinement delivers 57 total preference points over the original games across 150 paired conditions, giving a mean of 0.380. Its 93 selected revisions comprise 32 first-ranked, 31 second-ranked, and 30 third-ranked candidates. The remaining 57 conditions retain their original games. The outcome counts are 35 increases, 111 ties, and four decreases.

\paragraph{Systematic diagnostic feedback.}
The feedback experiment revisits 52 eligible retained-original conditions. In each condition, the source is the lowest numerical rank with valid gameplay and exact matching experience records. One feedback-driven rule revision preserves that source candidate's target and direction. The batch produces 50 valid scored candidates. Ten satisfy the deployed admission rule. Updating the selected outputs gives 103 adopted revisions and 61 total preference points, for a mean of $61/150=0.40\overline{6}$. The dedicated Crazy Eights case has a separate record and is excluded from these aggregates.

\paragraph{Sensitivity of the admission rule.}
Table~\ref{tab:gating} applies alternative admission rules to the fixed pool of delivered candidates: the 93 selected revisions and 50 valid feedback candidates. Every row uses the same executability requirement. This post-hoc analysis describes selection behavior on that pool. The deployed gate gives the main feedback mean of 0.407; the prediction-only rule selects 121 candidates with a mean of 0.480. These values expose a practical choice between requiring the intended target movement and accepting a broader set of predicted preference gains.

\begin{table}[t]
\centering\small
\setlength{\tabcolsep}{4pt}
\caption{\textbf{Sensitivity of refinement outcomes to the admission rule.} Alternative rules trade off the number of adopted revisions and observed preference changes. Italics identify the rule used in the refinement experiment.}
\label{tab:gating}
\begin{tabularx}{\linewidth}{Xcccc}
\toprule
Admission gate on the fixed pool & Adopted / 150 & From feedback & Mean $\Delta y$ & Imp. / Tie / Dec.\\
\midrule
\textit{Target $\geq0.25$ and prediction $>0$} & \textit{103} & \textit{10} & $\mathit{+0.407}$ & \textit{39 / 107 / 4}\\
Target $\geq0.25$ only & 106 & 13 & $+0.400$ & 40 / 104 / 6\\
Prediction $>0$ only & 121 & 28 & \textbf{$+0.480$} & 48 / 97 / 5\\
Target non-decreasing and prediction $>0$ & 115 & 22 & $+0.433$ & 44 / 101 / 5\\
Validity only & 143 & 50 & $+0.420$ & 56 / 82 / 12\\
\bottomrule
\end{tabularx}
\end{table}

\subsection{Case Record and Verification}
\label{app:case-evidence}
The Crazy Eights example is the {separate} feedback pilot for Thrill Seeker and the challenge-fit target. It is excluded from {both the systematic feedback aggregate and the method comparison, each covering 150 conditions}. We checked the revised rulebook, completed-play status, original experience responses, median preference record, saved gate decision, and frozen target coefficients. The core experience responses give challenge fit 3; the three separately collected preference evaluations have median 6. The saved baseline preference is 2. Summing the original-unit coefficient contributions reproduces the recorded predicted gain of 3.276. The gate record explicitly excludes observed preference and marks preference assessment as occurring after selection.

The first proposal reveals a player's hand after extended drawing. Its challenge-fit responses are all 2, and the saved critique emphasizes weak immediate consequences of drawing. The revised rule limits drawing to three cards, permits stopping after an unsuccessful draw, and buries the highest-value card on a bust. This {reduces both hand size and remaining hand value}.

\section{Implementation and Baseline Details}
\label{app:contracts}

\paragraph{Illustrative obligation.}
For a private-inspection rule, the observation function assumes that the intended recipient and card are recorded in the pending-inspection state. Its guarantee requires delivering the card to that recipient while preserving other players' information boundaries. The provider set identifies functions that can establish this pending state. The inspection event activates the obligation, and the recipient's next decision sets its deadline, expressing that delivery must occur in time to inform that decision. A function may own multiple clauses, each remaining linked to its supporting rule requirement.

\paragraph{Planning.}
A contract plan maps each rule requirement to its owner function and lists possible prerequisite providers. Every clause has a stable identity, executable assumption and guarantee predicates, and an obligation trigger and deadline. Structural checks and requirement coverage determine plan acceptance; semantic-review advice informs plan revision. The preparation stage permits four attempts and records semantic-review advice. Accepted predicates and clause identities remain frozen through the implementation loop.

\paragraph{Local execution.}
A fixture begins from a reference reset-state copy and constructs entry arguments satisfying the owner's assumption. The candidate owner executes on that input. The guarantee evaluates the actual before-state, after-state, and return value. Fixtures establish their prerequisites independently of the candidate provider functions, giving the local result a direct interpretation as an owner check.

\paragraph{Joint execution.}
The runtime observes real function-entry events and associates them with activated obligations. At a matching owner call, it applies the same assumption predicate used locally. A reached deadline with a missing required call is recorded as a failed obligation. Candidate providers are drawn from declared relationships and execution evidence in the obligation window. Check outcomes distinguish pass, fail, and unknown, according to the observed predicate and execution coverage.

\paragraph{Scope selection and repair.}
In Equation~\ref{eq:cover}, the cost of a function is its inclusive source-line count, computed as its end line minus its start line plus one. The search in Equation~\ref{eq:cover-recursion} explores candidates in ascending cost order. Among scopes with equal total cost, it prefers fewer functions and then breaks ties by sorted function names. Each search is bounded to two seconds; reaching this limit leaves scope selection unresolved and returns no supported repair scope. The responsibility sets contain candidate functions supported by the observed execution, and failures without an evidence-supported responsibility set remain unresolved. Source edits are restricted to selected existing function bodies with uniquely matched replacements; signatures, rule text, and checking predicates are preserved. Up to two repair rounds test improvement while preserving prior passing cases.

\paragraph{Baseline adaptations.}
\label{app:baseline-adaptations}
Cardiverse follows its rulebook-to-code workflow using implementation templates \citep{li2025cardiverse}. Reflexion adapts self-feedback refinement \citep{shinn2023reflexion}: feedback on the original persona play guides rule revision, and a reflection on the required changes precedes code editing and execution checks. ReCode adapts recursive code generation \citep{yu2025recode} with a plan limited to two levels and four executable leaf steps. Code is updated and tested after each step. PromptBreeder adapts prompt mutation \citep{fernando2024promptbreeder} using a fixed mutation prompt to generate one rule variant per game for all five personas. In the shared-rule implementation tasks, the same prompt supplies an implementation strategy for the prescribed target rulebook.

\paragraph{Implementation evaluation protocol.}
\label{app:implementation-protocol}
The implementation evaluation uses the shared target rulebooks described below. Preference refinement uses localized edits with execution checks; its playtesting and outcome aggregation are described in Appendix~\ref{app:representation} and Appendix~\ref{app:historical-refinement}.

The benchmark contains 30 game-level rule-revision tasks, with a frozen target assigned to each task and personas rotated across games. Each program is evaluated with a random policy over seeds 52000-52009. The PCon evaluator uses the final six decisions and the game outcome to assess agreement with the revised rulebook.

\paragraph{Outcome aggregation.}
Success rate counts final implementations that complete all ten evaluation games, with all 30 tasks as the denominator. For PCon, terminal non-delivery contributes zero; delivered outputs use valid paired seeds. We average within games and then weight games equally. This yields 29 comparable games and 286 paired seed positions. The implementation summary reports each system's final output.

\paragraph{Execution and embedding metrics.}
Exec replays each successful final implementation on 100 additional random seeds, 53000-53099, and averages its error-free fraction across games. Each rollout uses the same random-policy worker as the ten-seed evaluation, with a 4,000-turn limit and a 150-second timeout. ECon uses Qwen3-Embedding-0.6B to embed the complete final code and its shared structured target rulebook. Their cosine similarity is averaged over successful implementations.

\section{From Assessment to an Executable Design Objective}
\label{sec:discussion}

\paragraph{A shared vocabulary makes different priorities inspectable.}
The model benchmark shows that similar game rankings can coexist with different dimension associations and scoring levels. Persona references interpret these patterns; refinement uses conditional experience-preference models to choose targets. Held-out prediction connects these measurements to preference on unseen games.

\paragraph{A fixed target connects feedback to a mechanic.}
Keeping the persona, dimension, and direction fixed lets feedback address the same design objective. The Crazy Eights record illustrates how an unchanged measurement guides a new drawing mechanic. The {30-game} comparison tests delivered revisions across games. The results concern LLM personas playing card games; human-player studies can examine how the approach transfers to other players and settings.

\paragraph{Implementation quality supports rule refinement.}
The intended experience depends on the behavior players encounter. Function contracts specify how cooperating functions should realize a revised rule. The implementation evaluation reports more runnable outputs and higher play-rule consistency for \method{} than for Cardiverse.

\end{document}